\documentclass[11pt,a4paper]{article}
\usepackage[hyperref]{ranlp2025}
\usepackage{times}
\usepackage{latexsym}

\usepackage{microtype}

\usepackage{tabularx}
\usepackage{hyperref}

\aclfinalcopy 
\def\aclpaperid{192} 

\title{Synthetic vs. Gold: The Role of LLM Generated Labels and Data in Cyberbullying Detection}

\author{
 Arefeh Kazemi \and  Sri Balaaji Natarajan Kalaivendan \and Joachim Wagner  \\ \and {\bf Hamza Qadeer} \and
  {\bf Kanishk Verma} {\bf \and} {\bf Brian Davis} \\
School of Computing, ADAPT Centre,
Dublin City University,
Dublin, Ireland
\\
  \small{
    \textbf{Correspondence:} \href{mailto:firstname.lastname@adaptcentre.ie}{firstname.lastname@adaptcentre.ie}
  }
}

\date{}

\begin{document}
\maketitle

\begin{abstract}
Cyberbullying (CB) presents a pressing threat, especially to children, underscoring the urgent need for robust detection systems to ensure online safety. While large-scale datasets on online abuse exist, there remains a significant gap in labeled data that specifically reflects the language and communication styles used by children.
The acquisition of such data from vulnerable populations, such as children, is challenging due to ethical, legal and technical barriers. Moreover, the creation of these datasets relies heavily on human annotation, which not only strains resources but also raises significant concerns due to annotators’ exposure to harmful content. In this paper, we address these challenges by leveraging Large Language Models (LLMs) to generate synthetic data and labels.
Our experiments demonstrate that synthetic data enables BERT-based CB classifiers to achieve performance close to that of those trained on fully authentic datasets (75.8\% vs. 81.5\% accuracy). Additionally, LLMs can effectively label authentic yet unlabeled data, allowing BERT classifiers to attain a comparable performance level (79.1\% vs. 81.5\% accuracy). These results highlight the potential of LLMs as a scalable, ethical, and cost-effective solution for generating data for CB detection.

\end{abstract}

\section{Introduction}
The rapid proliferation of social media platforms has raised concerns over the prevalence of cyberbullying (CB), particularly among vulnerable populations (children). CB is defined as an aggressive behaviour directed towards someone with an intent to cause harm via electronic means \cite{patchin2006bullies, smith2008cyberbullying}.  
Detecting and mitigating CB is crucial to maintaining a safe digital environment and minimizing its psychological impact \cite{arif-etal-2024-impact}.
Despite the availability of large-scale online abuse datasets, there remains a notable lack of labelled data reflecting children's distinct linguistic and communicative styles. Collecting data from vulnerable groups, such as children, poses significant ethical, legal, and technical challenges. Moreover, creating such datasets is hindered by high costs, time-consuming annotation processes, and ethical risks, including exposure to distressing content or potential re-traumatization of annotators \cite{alemadi-Zaghouani-2024-emotional}. Across a variety of tasks, large language models (LLMs), particularly instruct-tuned variants, offer a promising approach to synthetically generate labeled data without requiring direct human exposure to harmful material.


\paragraph{Industry Use Case:}
This paper is motivated by an nationally funded industry/academic research collaboration whereby the lead industry partner is a European SME is developing integrated child-protection software into children's smartphones to ensure safer internet use. By analyzing incoming and outgoing data on a child's device using AI/NLP techniques, the software filters offensive content, including CB, and alerts parents if any imminent danger is detected.   Hence, our are uses case is constrained by i) the need for \textit{age relevant} CB test/training  data and ii)  using an LLM directly as classifier is too expensive due to the high volume of messages to be checked on the smart device.
However, collecting authentic CB content produced by children is difficult. Age information on public social media sites is unreliable and a large part of CB happens in closed groups. Furthermore, obtaining consent involves the parents, who may not be interested in (or may not have time to consider) taking part in a research study.
A solution may be to use LLMs to generate conversations between
children.


In this paper, we explore the potential of LLM-generated datasets for CB detection. We investigate several scenarios for integrating LLMs into the CB detection task. Through these experiments, we aim to determine the extent to which synthetic data can improve CB detection performance, considering different levels of availability of manually labeled authentic data. All synthetic datasets, along with the synthetic labels generated for the authentic dataset, will be made available upon request for research purposes.

\section{Background}
Recently, LLMs have emerged as efficient, scalable, and cost-effective alternatives to manual annotation for synthetic data generation.
\newcite{he-etal-2021-generate, he-etal-2022-generate} leveraged LLMs for generating data in knowledge distillation, self-training, and few-shot learning, later labeled by state-of-the-art sequence classifiers. 
\newcite{bonifacio-etal-2022-inpars} employed LLMs
to generate labeled data in a few-shot manner for information retrieval.
\newcite{yoo-etal-2021-gpt3mix} created augmented samples by embedding training sentences into prompts. 
\newcite{anaby-etal-2020-donot} fine-tuned a language model on limited labeled data, then used it to produce labeled text. 
\newcite{meng-etal-2022-generating} employed an LLM
to generate class-conditioned texts based on label-descriptive prompts for a classification task. Moreover, research has explored 
synthetic data generation for specific domains, such as medical and harmful content detection.
\newcite{wang-etal-2024-notechat} introduced 
a multi-agent framework designed to generate synthetic patient-physician conversations from clinical documents. \newcite{ghanadian-etal-2024-socially} created socially aware synthetic datasets for suicidal ideation detection.
Previous studies exploring the use of synthetic data have reported mixed results on whether LLM-generated synthetic data can effectively train models to perform at a level comparable to those trained on authentic data \cite{li-etal-2023-synthetic}.

Additionally, \newcite{ejaz-etal-2024-multi} constructed a 
CB dataset by simulating user interactions, where aggressive and non-aggressive messages were drawn from authentic datasets and randomly assigned to ``synthetic'' users. They annotated CB instances using a threshold-based method, considering their own criterion like ``peerness'', ``harmful intent'', and ``repetition''. 
\newcite{perez-etal-2024-generation} 
created a synthetic dataset of CB risk scenarios for serious games using a Bayesian Network model \cite{pearl-1988-probabilistic} trained on survey data from minors. Data were generated via Bayesian Model Sampling, yielding agent profiles with demographics and responses to 15 binary CB-related questions. This dataset does not include actual CB messages; instead, it represents scenarios using structured, non-textual features.
Moreover, recent work 
explores the use of LLMs for synthetic data generation and augmentation to address data scarcity:
\newcite{electronics13173431} addresses the intersection of CB and bias detection. They use LLMs both for generating and detecting biased and abusive content.
\newcite{hui-etal-2024-toxicraft} use a small amount of seed data to generate synthetic examples of harmful content. However, their approach still relies on seed data and does not specifically address the nuanced communication patterns characteristic of young people.
\newcite{schmidhuber-kruschwitz-2024-llm} test data augmentation with synthetic data in the task of toxic language detection.
For each authentic dataset, they fine-tune two GPT-3 Curie 6.7B \cite{brown-etal-2020-language} models, one on toxic language and one on non-toxic language.
Then, synthetic data is generated for each toxicity class, filtered for quality and used to train classifiers.
They experiment with six authentic datasets, including one none-English dataset, a German dataset.

In contrast to these works, our study aims to explore the use of LLMs to generate a fully synthetic CB dataset or to label existing authentic but unlabeled CB datasets.
Our work uniquely applies LLM-generated data to age relevant Cyberbullying(CB) detection, an area where data collection is both ethically and logistically challenging.
By assessing the effectiveness of LLM-generated datasets in comparison to authentic data,
we provide new insights into the viability of LLMs in training models for sensitive tasks like CB detection.

\section{Methodology}\label{s:methodology}
\subsection{Overview of Scenarios}

We investigate the role of LLMs in CB detection, focusing on their utility under varying data availability conditions since our industry use case indicates that direct use of LLMs as a classifier is too expensive due to the high volume of messages to be checked.
As a baseline for comparison, we
evaluate a scenario in which a
lightweight, BERT-based
classifier
is trained exclusively on gold-standard, manually labeled authentic data without
LLM involvement.
We then define three additional scenarios with different data availability
and that use LLMs in different ways.

\paragraph{Scenario 1: Baseline}

This scenario represents the ideal situation where sufficient
gold-standard data is available for fine-tuning a classic encoder such as
BERT \cite{devlin-etal-2019-bert}. It serves as the benchmark for evaluating the effectiveness of other approaches.
In this setup, no synthetic data or LLMs are involved.
This scenario is feasible if resources such as time, budget, and expert annotators are abundant. However, it often proves impractical due to the
challenges of manual labeling.

\paragraph{Scenario 2: LLM as Classifier}  \label{s:m:sc2}

This scenario applies when labeled authentic data is unavailable, and there is no intention to train a classifier for CB detection. Instead, an instruction-tuned
LLM is used directly as a classifier, leveraging its pre-trained knowledge and its ability to follow instructions
to identify CB instances.
The primary advantage of this method is its elimination of the need for labeled data and training time. However, there are trade-offs. While an LLM can handle nuanced language patterns, it may be less efficient and incur higher computational costs
compared to simpler BERT-based classifiers with a classification head and fine-tuned on a labeled dataset.
We explore two prompting strategies for generating synthetic labels:
\textit{(a)} guideline-enriched (GE) prompts, guiding the LLM with detailed labeling instructions, and
\textit{(b)} guideline-free (GF) prompts, allowing the LLM to generate labels without such guidelines.

\paragraph{Scenario 3: Fully Synthetic Data}

In this scenario, only a small set of manually labeled gold data is available for testing, with no access to authentic data for training or validation.
We
use an LLM to generate a fully synthetic dataset, consisting of both synthetic messages and corresponding labels, for training and validation.
This approach is particularly valuable in low-resource domains or emerging tasks where authentic data is scarce or difficult to collect.
It is especially useful in situations where creating authentic datasets is costly, time-consuming, or ethically challenging, such as annotating harmful or sensitive content or working with vulnerable populations.

\paragraph{Scenario 4: Synthetic Labels for Unlabeled Data} \label{s:m:sc4}

This scenario addresses the common situation where resources for manual annotation are limited. Here, gold-standard labeled data is available only for the test set, while a significant amount of unlabeled authentic data is available for training and validation.
To utilize the unlabeled data, we label it using the best prompting strategy (GE or GF) from scenario~2.

\subsection{Evaluation Metrics}

We choose accuracy\footnote{In the appendix, we further report macro average F1 scores that are also widely used in harmful content detection. 
}
of label prediction for development decisions and reporting since the labels are reasonably balanced in the authentic test data, with 30.3\% items labeled with the minority
label.\footnote{The
   label distribution reflects the construction of the dateset via role-playing, see Section~\ref{s:setup:dataset}.
}
In scenarios 1, 3 and 4,
we train BERT\_base\_uncased \cite{devlin-etal-2019-bert}, a 110M parameter transformer model, with a linear classification head\footnote{Using defaults of BERT sequence classification with
the HuggingFace transformers library \cite{wolf-etal-2020-huggingface}}
to detect harm, assigning binary labels
to text messages.\footnote{We
    choose this model as it has been adopted widely and performs
    competitively for sequence classification tasks
    \cite{galke2022we}.
    Although more modern architectures have been proposed, BERT's
    well-tested performance and deep understanding within the research
    community make it a reliable choice
    for our comparative evaluations. 
}
To address noise from randomness in training, we train at least
45 repetitions with different random seeds\footnote{Initial
    experiments showed unstable results when moving from 25 to 35 repetitions.
    We speculate that the small size of the dataset and
    the subjectivity of the task cause the noise.}
for each setting and report average accuracy and standard deviation.

\section{Experimental Setup}
\subsection{Dataset Description}
\label{s:setup:dataset}

The primary objective of our study is to detect cyberbullying (CB) among children aged 6 to 16 years.
Therefore, it is essential to utilize a dataset specifically tailored for this purpose.
Focused on pre-adolescents, we leverage a well known CB  dataset for young demographics, which was constructed through role-play activities in WhatsApp (WA) groups, each containing approximately 10 Italian students  ~\cite{sprugnoli-etal-2018-creating}. We use the English version of this dataset for our experiments.
The roles are:
    cyberbully (2 students),
    cyberbully assistants (3–4 students),
    victim assistants (3–4 students) and
    a victim.
Conversations are initiated using one of four
predefined cases (A to D) \cite{sprugnoli-etal-2018-creating} describing situations in which CB may occur. The four cases illustrate some of the situations in which cyberbullying may occur among students. Case A addresses the ``gendered division of sport practices,'' where a shy male student defies gender norms by inviting his classmates to watch his ballet performance. Case B involves ``interference in others' businesses,'' as a high-achieving student reports peers for bringing cigarettes to school, resulting in disciplinary action and social exclusion. Case C highlights ``lack of independence, parental intromission,'' where a student's parent convinces teachers to assign more homework, causing resentment among classmates. Case D deals with ``web virality,'' where a shy student becomes the subject of mockery after a video of them dancing awkwardly surfaces online. Table~\ref{t:whatsapp-scenarios} shows one of these cases as an
example.\footnote{In
    the cited paper, these cases are referred to
    as ``scenarios.''
    However, we use the term ``cases'' to avoid confusion with
    the scenarios discussed above.
}

\begin{table}[ht]
\centering
\begin{tabularx}{\columnwidth}{XX}
\textbf{Case} & \textbf{Type of Problem}\\
\hline
\textbf{A:}
Your shy male classmate has a great passion for classical dance.
Usually he does not talk much, but today he has decided to invite the class to watch him for his ballet show. &

Gendered division of sport practices \\
\hline
    \end{tabularx}
    \caption{Example case and problem type for conversation initiation~\cite{sprugnoli-etal-2018-creating}}
    \label{t:whatsapp-scenarios} 
\end{table}

Messages in the conversation are annotated manually by experts following the fine-grained framework
of \newcite{van-hee-et-al-2015-guidelines}. The annotations categorized CB instances into detailed types, such as General Insults, Body Shame, Sexism, and Racism.
However, for this study, we binarize the labels by mapping all categories of CB by bullies to the ``harmful''
label\footnote{We
    deviate from \newcite{verma-etal-2023-leveraging} in not excluding harmful messages of the victim from the ``harmful'' class.
}
Table~\ref{t:dataset-stat} shows the statistics of the WA dataset.

\begin{table}
\centering
\begin{tabular}{lrrc}
\textbf{Split}  & \textbf{Size} & \textbf{Harmful}  & \textbf{\%} \\
\hline     
Training        &  1,314        &   398    &  30.3 \\ 
Validation      &    439        &   133    &  30.3 \\ 
Test            &    439        &   133    &  30.3 \\     
\hline
Total           &  2,192        &   664    &  30.3 \\ 
\hline
\end{tabular}
\caption{WA dataset: number of messages and fraction of harmful messages}
\label{t:dataset-stat}
\end{table}

\subsection{Generating Synthetic Data and Labels}

We employ three LLMs for our experiments:
GPT-4o-Feb-2025 \cite{openai-2024-gpt},
Llama-3.3-70B-Instruct \cite{meta-2024-llama33}
\footnote{We include a small amount of synthetic data produced by Llama3 from early experiments, see Appendix Table ~\ref{t:llama-versions}. For labeling with Llama, version 3.3 has been used throughout.
} and Grok-2-Feb-2025 \cite{xai-2024-grok}.  Our prompt design process for all scenarios begins with a simple initial prompt, which is 
refined over multiple rounds of trial and error.
In this iterative approach, we make gradual improvements, with adjustments made based on the quality and relevance of the responses generated by the LLM on the development set.

\subsubsection{Designing Prompts for Synthetic Label Generation}

In the task of labeling the messages of a conversation as either harmful or not, the input data consists of a list of messages and the possible labels are ``Harm'' and
``No Harm''.\footnote{We do not provide the LLM with instructions on how to handle victim defense and other special cases.}
We simplify the task by not considering context (previous messages) and classifying each message in
isolation.\footnote{Future work can consider how to best include the conversation context into the prompt.}
As described in Section \ref{s:methodology} We explore two approaches to prompt design: (1) guideline-free (GF) and  (2) guideline-enriched (GE).
In the GF approach, the LLM is simply instructed to label messages as ``Harm'' or ``No Harm'' for the task of CB detection,
without providing additional guidelines.
In the GE approach, the LLM is supplied with detailed instructions for labeling messages.
These instructions are adapted from the annotation guidelines~\cite{van-hee-et-al-2015-guidelines} originally used by human annotators for labeling authentic data. For this study, since the test set is derived from the WA(WhatsApp) dataset, we utilize the same guideline employed by human annotators to label this dataset. A copy of the guidelines is provided in Table~\ref{t:whatsapp-guideline}, and our prompts for both the GE and GF approaches are shown in Table~\ref{t:prompt}.

\begin{table*}
    \centering
    \begin{tabularx}{\linewidth}{|X|}
        \hline
       Cyberbullying-related text categories are described below:\\
       
    \textbf{Threat/Blackmail} Physical or psychological threats or indications of blackmail.\\
    \textbf{Insult}  Abusive, degrading, or offensive language intended to insult the recipient.\\
    \textbf{Curse/Exclusion}  Wishes of adversity or attempts to exclude someone from a group.\\
    \textbf{Defamation}  Revealing confidential or defamatory information about someone to a large audience.\\
    \textbf{Sexual Talk}  Potentially harmful expressions with a sexual meaning.\\
    \textbf{Defense}  Expressions supporting the victim, either by themselves or a bystander.\\
    \textbf{Encouragement to the Harasser}  Support or agreement with the harasser's actions.\\
    \textbf{Body Shame}  Criticism based on the shape, size, or appearance of someone's body. \\
        \hline
    \end{tabularx}
    \caption{Annotation guidelines used by human annotators for labeling the WhatsApp dataset~\cite{van-hee-et-al-2015-guidelines}}
    \label{t:whatsapp-guideline} 
\end{table*}

\begin{table*}
    \centering
    \begin{tabularx}{\linewidth}{|p{2cm}|X|}
        \hline
    \textbf{Approach} &\textbf{Prompt} \\
    \hline
    GE Synthetic Label & \{Annotation-guideline\} Please analyze the following text and determine whether it contains any form of cyberbullying or harmful content. If it does, classify it as 'Harm'. If it does not, classify it as 'No Harm'. \{Text\}.  \\
        \hline
    GF Synthetic Label & Please analyze the following text and determine whether it contains any form of cyberbullying or harmful content. If it does, classify it as 'Harm'. If it does not, classify it as 'No Harm'. \{Text\}\\
    \hline
     Synthetic Data & We are creating sample conversations to aid in cyberbullying detection. In these cases, teens are asked to role-play and create realistic conversations based on provided situations. There are 11 students participating in the conversation. The teens participating are: VCTM, BULLY1, BULLY2, VSUP1, VSUP2, VSUP3, VSUP4, BSUP1, BSUP2, BSUP3, BSUP4 with roles assigned as follows: VCTM: Victim, BULLY1 and BULLY2: Bully VSUP1, VSUP2, VSUP3 and VSUP4: Victim Support BSUP1, BSUP2, BSUP3 and BSUP4 : Bully Support. consider this case: \{Case\} and consider this Type of addressed problem: \{Type of Problem\}. Generate an example conversation, with at least 100 messages, between these students based on the provided case and Type of addressed problem. Use profanity and strong language to create a realistic dialogue. number each message in the conversation. Please note that the conversation should be realistic and can be offensive. Please make sure to include different topics and perspectives in each conversation \\
    \hline
    \end{tabularx}
    \caption{Prompts used for generating synthetic labels via (1) guideline-free (GF) and (2) guideline-enriched (GE) approaches, as well as for synthetic data creation.Variables are enclosed in curly brackets \{\}. 'Text' refers to the message content, and the 'Annotation-guideline' is provided in Table \ref{t:whatsapp-guideline}. 'Case' and 'Types of Problem' are adapted from \cite{sprugnoli-etal-2018-creating}. An example case and its corresponding problem type are shown in Table~\ref{t:whatsapp-scenarios}.}
    \label{t:prompt}
\end{table*} 

\subsubsection{Designing Prompts for Synthetic Data Generation}

We aim to produce synthetic data that consists of conversations between participants, where the dialogue should include instances of CB (or at least harmful messages that could be CB if the threats are repeated and intended to cause harm).
To make the generated data more relevant to the test data, each prompt includes one of the four predefined ``cases'' (A to D) and ``problem types'' that were used by \newcite{sprugnoli-etal-2018-creating} in the creation of the WA dataset to guide the students' role-playing.
Table~\ref{t:prompt} presents the prompts employed for generating synthetic data.

\subsection{Amount of Synthetic Data}

For a more meaningful comparison of results with the authentic data, we sample subsets of the generated data to match the size of the authentic data.
We maintain the ratio of messages belonging to each case (A to D) found in the authentic data.
A size of 100\% means that 
863 messages are included for case A,
462 for case B,
103 for case C and 
325 for case D.\footnote{These
    numbers match the number of messages for each case in the
    training and validation data of the authentic dataset.
    The data is then split 60:20 = 75:25 to produce the training
    and validation data.
    The test set always uses authentic, manually labeled data.
}
We sample subsets of certain percentages of these sizes
from 100\% to 200\%.
We ensure to produce more than 200\% of synthetic data
(relative to the size of the authentic data) for each
LLM and each case so that these samples can be taken
via sampling without replacement, not duplicating any
data.\footnote{The smallest available amount is approximately 208\% for case B with Grok.}

%
%

\subsection{Authentic Unlabelled Data}

In \textbf{Scenario~4}, we remove the gold labels from the training and validation section of the authentic labelled data of \textbf{Scenario 1} to produce the unlabelled data
to maximise comparability of the data and to limit changes to the labels that are
annotated by the LLM in this scenario.
In both  \textbf{Scenarios 3 and 4}, the validation set with synthetic labels is used for model selection and early stopping, but validation set results in Section~\ref{s:results} will use the gold labels
to measure performance.\footnote{Accuracy
   on synthetic validation data often exceeds 90\% in our experiments.
   This may mean that the synthetic data is less diverse than the authentic data for which validation accuracy usually is below 83\%, or that harm is more clearly
   expressed.
}

\section{Results and Discussion} \label{s:results}



\subsection{Scenario 1: Baseline}

\begin{table}
\centering
\begin{tabular}{rcc}
\textbf{Size}  & \textbf{Dev.} & \textbf{Test}  \\
\hline
20\%  &   74.2\% $\pm$    2.8  &   73.7\% $\pm$    2.7  \\
50\%  &   78.8\% $\pm$    2.1  &   79.4\% $\pm$    1.7  \\
80\%  &   80.0\% $\pm$    1.8  &   80.4\% $\pm$    1.5  \\
100\%  &   80.9\% $\pm$    1.6  &   81.5\% $\pm$    1.2  \\
\hline
\end{tabular}
\caption{Development and test set accuracy in scenario~1: training BERT-based classifiers on the training split of the
authentic data;
at least 
45 repetitions with different random seeds;
also shown for comparison results for training on samples
of 20\%, 80\% and 50\%
(sampling without replacement;
both the training set and the development set are sampled
to the given relative size of the authentic data split). Macro-F1 scores and more detailed results are available in Table~\ref{t:results-s1-a} in the Appendix.
}
\label{t:results-s1-b}
\end{table}

Table~\ref{t:results-s1-b} shows the development and test set accuracies for BERT-based classifier trained on different portions of the training split of authentic data in \textbf{Scenario~1}.
The results illustrate the impact of training data size on the performance of a BERT-based classifier.
As expected, model accuracy on both the development and test sets improves as the portion of authentic training data increases. The improvement is most pronounced between 20\% and 50\% of the data, where test accuracy jumps by nearly 6\%, indicating that a significant portion of the model's learning occurs with moderate data availability. However, gains diminish beyond 80\%, with only a marginal increase of 1.1 percentage points when scaling from 80\% to 100\%. This suggests diminishing returns in model performance as more data is added.

\subsection{Scenario 2: LLMs as a Classifier}

Table~\ref{t:results-s2} presents the accuracy of the Llama 3.3 70B, GPT-4o, and Grok-2-1212 models using guideline-enriched (GE) and guideline-free (GF) prompts on both the development and test sets in \textbf{Scenario~2}. We evaluate both GE and GF prompts on the development set and test the winning prompts on the test set.
GPT-4o achieves the highest accuracy regardless of the prompt used, 
while Grok performs the lowest.
The lower performance of GF prompts for two LLMs, particularly for Grok, highlights the importance of prompt design for model accuracy.

To decide which prompt to evaluate on test data for GPT-4o, we use the consistent performance trend with the other LLMs as a tie breaker.
Comparing the guideline-enriched (GE)test set results in Table~\ref{t:results-s2} with those in
Table~\ref{t:results-s1-b}, GPT-4o with the GE prompt reaches the highest test accuracy, outperforming the BERT-based model by 1.6\%.
However, despite the performance advantage, LLMs come with significant computational costs, making them less practical for real-time applications on resource-constrained devices.
In contrast, the BERT-based classifier is a lightweight model optimized for efficiency, making it a more suitable choice for scenarios where speed and scalability are crucial. Since our industry partner aims to deploy CB detection on children's mobile devices, we prefer using the BERT-based classifier due to its lightweight nature, which allows it to run efficiently on mobile hardware.
This trade-off highlights the importance of balancing accuracy with
application requirements, as a small accuracy decrease of
1.6\% can be acceptable in exchange for a model that is faster, more accessible, and capable of running on mobile devices.

\begin{table}[ht]
\centering
\begin{tabular}{llrr}
\textbf{LLM} & \textbf{Prompt} & \textbf{Dev. Set} & \textbf{Test Set} \\
\hline
Llama & GE & 82.9\% & 80.4\% \\
Llama & GF & 82.0\% & -\\
\hline
GPT & GE & \textbf{85.6\%} & \textbf{83.1\%} \\
GPT & GF & \textbf{85.6\%} & - \\
\hline
Grok & GE & 81.3\% & 78.6\% \\
Grok & GF & 75.9\% & - \\
\hline
\end{tabular}
\caption{Accuracy of various LLMs using GE and GF prompts on the development set, and using the winning prompt (always GE) on the test set in Scenario 2: LLM as a CB classifier.}
\label{t:results-s2}
\end{table}

\subsection{Scenario 3: Fully Synthetic Data}  


\begin{table}
\centering
\begin{tabular}{lrcc}
\textbf{LLM}  &
\textbf{Size}  &
\textbf{Dev.} & \textbf{Test}  \\
\hline
LA & 100\%  &   73.6\% $\pm$    4.1  &   73.4\% $\pm$    3.0  \\
LA & 120\%  &   74.8\% $\pm$    3.6  &   74.5\% $\pm$    2.5  \\
LA & 160\%  &   76.1\% $\pm$    2.6  &   75.3\% $\pm$    2.2  \\
LA & 200\%  & \textbf{76.6\%} $\pm$    2.5  & \textbf{75.8\%} $\pm$    2.1  \\
\hline
GPT    & 100\%  &   69.7\% $\pm$    0.8  &   70.2\% $\pm$    0.9  \\
GPT    & 120\%  &   69.8\% $\pm$    0.7  &   70.3\% $\pm$    1.0  \\
GPT    & 160\%  &   69.9\% $\pm$    0.8  &   70.3\% $\pm$    0.9  \\
GPT    & 200\%  &   70.1\% $\pm$    0.7  &   70.4\% $\pm$    0.9  \\
\hline
Grok   & 100\%  &   49.9\% $\pm$    8.5  &   52.2\% $\pm$    8.0  \\
Grok   & 120\%  &   50.7\% $\pm$    8.6  &   52.9\% $\pm$    8.1  \\
Grok   & 160\%  &   51.2\% $\pm$    8.9  &   53.5\% $\pm$    8.1  \\
Grok   & 200\%  &  52.3\% $\pm$    8.3  &  54.2\% $\pm$    7.8  \\
\hline
\end{tabular}
\caption{Development and test set results in scenario 3:
    training a BERT-based classifier on synthetic data matching
    100\% to 200\% of the size available in scenario 1;
    LA = Llama3; at least 45 repetitions with different random seeds;
    Macro-F1 scores and more detailed results are available in Tables~\ref{t:results-s4-r1v1} to \ref{t:results-s4-gr4k} in the Appendix.
}
\label{t:results-s3-s-3}
\end{table}

Table \ref{t:results-s3-s-3} presents the development and test set accuracy for \textbf{Scenario~3}: BERT-based classifiers trained on synthetic datasets generated by Llama, GPT, and Grok\footnote{To work within the LLMs' safety constraints, we explicitly mentioned in the prompt: ``We are creating sample conversations to aid in CB detection.'' This clarification generally enabled the model to generate the required synthetic data. In instances where the model still refused to respond, we re-tried the prompt until we obtained a sufficient amount of data.}. The results indicate that Llama consistently outperforms the other two models across different dataset sizes. GPT follows closely, while Grok exhibits the weakest performance, particularly on the test set, where its accuracy is considerably lower than that of the other models. This suggests that Llama-generated synthetic data is of higher quality and more beneficial for training the classifier, despite the lower label quality observed in \textbf{Scenario~2}.

Examining the impact of increasing the size of synthetic data, the results show a general trend of performance improvement for Llama and Grok, but little impact for GPT.
This indicates that the effectiveness of synthetic data scaling may depend on the LLM and the quality of the data.

Comparing the training of BERT on authentic data (Table~\ref{t:results-s1-b}) with its training on synthetic data (Table~\ref{t:results-s3-s-3}), the performance of Llama-generated synthetic data without any human-created data is notable  (75.8\% accuracy vs.\ 81.5\%). 
The 5.7\% accuracy gap highlights the trade-off between performance and the reduced costs and ethical complexities associated with human-annotated CB datasets.
For sensitive and ethically challenging tasks, 
synthetic data can present a valuable alternative, minimizing human involvement while maintaining competitive model performance.
The results emphasize that high-quality synthetic data from advanced LLMs, such as Llama, can serve as a practical supplement or even a partial substitute, particularly in scenarios where authentic data collection is difficult or impractical.

\subsection{Scenario 4: Synthetic Labels for Unlabeled Authentic Data}

\begin{table}
\centering
\begin{tabular}{lrcc}
\textbf{Labels} & \textbf{Size}  & \textbf{Dev.} & \textbf{Test}  \\
\hline
LA  &   20\%  &   74.8\% $\pm$    1.9  &   73.7\% $\pm$    1.7  \\
LA  &   50\%  &   79.3\% $\pm$    1.5  &   78.0\% $\pm$    1.3  \\
LA  &   80\%  &   80.1\% $\pm$    0.9  &   78.8\% $\pm$    1.0  \\
LA  &   100\%  &   \textbf{80.4\%} $\pm$    1.0  &   \textbf{79.1\%} $\pm$    1.0  \\
\hline
GPT  &   20\%  &   72.0\% $\pm$    1.7  &   72.6\% $\pm$    1.6  \\
GPT  &   50\%  &   75.1\% $\pm$    1.7  &   76.4\% $\pm$    1.3  \\
GPT  &   80\%  &   76.2\% $\pm$    1.1  &   77.2\% $\pm$    0.9  \\
GPT  &   100\%  &   77.1\% $\pm$    0.9  &   77.9\% $\pm$    0.9  \\
\hline
Grok  &   20\%  &   71.3\% $\pm$    2.6  &   70.4\% $\pm$    3.0  \\
Grok  &   50\%  &   75.0\% $\pm$    1.6  &   74.2\% $\pm$    2.0  \\
Grok  &   80\%  &   76.1\% $\pm$    1.1  &   74.4\% $\pm$    1.6  \\
Grok  &   100\%  &   76.3\% $\pm$    1.2  &   75.2\% $\pm$    1.7  \\
\hline
\end{tabular}
\caption{Development and test set accuracy in scenario~4: training BERT-based classifiers on the training split of the
authentic data with synthetic labels predicted by
(a) LA = Llama 3.3, removing dataset items for which no label is found in the LLM output,
(b) GPT = GPT-4o and
(c) Grok; at least 45 repetitions with different random seeds;
also shown for comparison results for training on samples
of 20\%, 80\% and 50\%
(sampling without replacement;
both the training set and the development set are sampled
to the given relative size of the authentic data split).
Macro-F1 scores and more detailed results are available in
Table~\ref{t:results-s4-a} in the Appendix.
}
\label{t:results-s4-c}
\end{table}

Table~\ref{t:results-s4-c} shows that among the different LLMs used for synthetic annotation, Llama 3.3 produces the most useful labels, allowing BERT to achieve a test accuracy of 79.1\%, closely approaching the human-labeled benchmark.
This is despite the fact that the results in \textbf{Scenario 2} above show that the GPT-4o labels better match human labels than Llama 3.3's labels.
In \textbf{Scenario 4},
GPT-4o follows with a slightly lower accuracy of 77.9\%, while Grok lags behind at 75.2\%. The difference in performance suggests that not all synthetic labels are of equal quality, and the choice of LLM for data annotation can impact model performance. Comparing the results in Table \ref{t:results-s1-b} with those in Table \ref{t:results-s1-a} shows that BERT trained on 100\% authentic human-labeled data achieves a test accuracy of 81.5\%, while the best-performing synthetic labels, produced by Llama 3.3, yield 79.1\%. This small gap of just 2.4\% highlights that synthetic labels can
be
a viable substitute
to human labels
while significantly reducing the cost, ethical concerns, and legal restrictions associated with human annotation, particularly for sensitive or harmful content.


\section{Conclusions}
In this paper, we explored the potential of LLM-generated synthetic data for cyberbullying detection, evaluating its effectiveness in \textbf{four} scenarios. 
Our results highlight that synthetic data can significantly reduce the reliance on human annotators while maintaining competitive model performance, especially in low-resource settings. However, we also observed that the quality and utility of synthetic data depend heavily on prompt design and data selection strategies.

For future work, we want to expand the set of data availability scenarios to include data augmentation, where authentic and synthetic data are combined.
Furthermore, we plan to work with social scientists working in online safety
to plan a human evaluation of the synthetic data.
We also plan to investigate how the temperature settings during label and conversation generation affect data diversity and model performance.

Future work may also want to more systematically tune the prompts than we did.
Our results may under-represent the full potential of LLMs to perform as automatic labelers and content creators.

Additionally, we want to try removing messages that are labeled with a low confidence score from the training data, similarly to, e.g., self-training \cite{yarowsky-1995-unsupervised,blum-mitchell-1998-combining}.
Finally, we intend to explore
selecting subsets of synthetic data based on their similarity to authentic data
for better alignment with real-world patterns.



\section*{Acknowledgments}  

This research is supported by the
Disruptive
Technologies Innovation Fund (DTIF) under the
project ``Cilter: Protecting Children Online'' Grant
No.\ DT~2021~0362 from the Department of Enterprise,
Trade and Employment in Ireland and administered by
Enterprise Ireland (EI).
This research is supported by the Research Ireland\footnote{Previously Irish Research Council}
Enterprise Partnership Scheme (EPS) with Google for Online Content Safety under grant number EPSPG/2021/161.
%
%
This research was conducted with the financial support of
Science Foundation Ireland under Grant Agreement
No.\ 13/RC/2106\_P2 at the ADAPT SFI Research Centre at
Dublin City University.
ADAPT, the SFI Research Centre for AI-Driven Digital Content Technology,
is funded by Science Foundation Ireland through the
SFI Research Centres Programme.
For the purpose of Open Access, the author has applied a
CC BY public copyright licence to any Author Accepted Manuscript version
arising from this submission.

\bibliographystyle{acl_natbib}  

\bibliography{main}




\appendix

\section{Appendix}\label{s:appendix}
\label{sec:appendix}

\begin{table}[ht]
\centering
\begin{tabular}{lrrrr}
\textbf{Case} & \textbf{3} & \textbf{3.3} & \textbf{Total} & \textbf{\% (3/Total)} \\
\hline
A     &   258 &  3,266 &  3,524 &   7.3 \\
B     &   332 &  1,648 &  1,980 &  16.8 \\
C     &   282 &  1,362 &  1,644 &  17.2 \\ 
D     &   100 &  1,012 &  1,112 &   9.0 \\ 
\hline
Total &   972 &  7,288 &  8,260 &  11.8 \\
\hline
\end{tabular}
\caption{Number of messages generated with each version of Llama used, broken down by case}
\label{t:llama-versions}
\end{table}

\begin{table*}[ht]
\centering
\caption{Advantages and Disadvantages of Each Scenario in Cyberbullying Detection}
\label{t:scenario-advantages-disadvantages}
\begin{tabularx}{\linewidth}{|X|X|X|}
\hline
\textbf{Scenario} & \textbf{Advantages} & \textbf{Disadvantages} \\ \hline

\textbf{Baseline (Gold-Standard Only)} & 
High-quality, reliable data &
High costs and scalability challenges; Requires significant time and expert annotation effort. \\ \hline

\textbf{LLM as Classifier} & 
No need for labeled data or training; Quick deployment; Handles nuanced language patterns. & 
Computationally expensive; May be less accurate than fine-tuned classifiers on domain-specific data. \\ \hline

\textbf{Synthetic Labels for Unlabeled Data} & Utilizes existing unlabeled data; Cost-effective dataset creation. & Label quality depends on LLM performance; May require validation to ensure consistency and accuracy. \\ \hline

\textbf{Fully Synthetic Data} & 
Enables training when no authentic data is available; Suitable for low-resource domains. & 
Synthetic data may lack diversity and realism; Risk of overfitting to generated patterns. \\ \hline



\end{tabularx}
\end{table*}


\begin{table*}[ht]
\centering
\begin{tabular}{rlccccr}
  &  & \multicolumn{2}{c}{\textbf{Development Set}} & \multicolumn{2}{c}{\textbf{Test Set}} \\
\textbf{Size} & \textbf{Sampling} & \textbf{Accuracy} & \textbf{Macro-F1} & \textbf{Accuracy} & \textbf{Macro-F1} & \textbf{Rep.} \\
\hline
100\%  &  up  &   79.8\% $\pm$    1.5  &   76.3\% $\pm$    1.8  &   80.8\% $\pm$    1.5  &   77.3\% $\pm$    1.5  & 50  \\
\hline
20\%  &  none  &   74.2\% $\pm$    2.8  &   68.1\% $\pm$    3.4  &   73.7\% $\pm$    2.7  &   67.7\% $\pm$    2.5  & 65  \\
50\%  &  none  &   78.8\% $\pm$    2.1  &   74.0\% $\pm$    2.6  &   79.4\% $\pm$    1.7  &   74.5\% $\pm$    2.1  & 65  \\
80\%  &  none  &   80.0\% $\pm$    1.8  &   75.6\% $\pm$    2.2  &   80.4\% $\pm$    1.5  &   75.8\% $\pm$    1.7  & 65  \\
100\%  &  none  &   80.9\% $\pm$    1.6  &   76.8\% $\pm$    1.8  &   81.5\% $\pm$    1.2  &   77.0\% $\pm$    1.6  & 85  \\
200\%  &  none  &   80.8\% $\pm$    1.6  &   76.6\% $\pm$    2.0  &   81.7\% $\pm$    1.2  &   77.2\% $\pm$    1.4  & 50  \\
\hline
\end{tabular}
\caption{Development and test set results in scenario~1: training BERT-based classifiers on the training split of the
authentic data; at least
45 repetitions with different random seeds;
also shown for comparison results for training on samples
from 20\% to 80\%, as well as two copies (200\%) of the data;
both the training set and the development set are sampled
to the given relative size of the authentic data split;
``Sampling'' refers to the strategy for addressing class imbalance in the training data}
\label{t:results-s1-a}
\end{table*}

\begin{table*}[ht]
\centering
\begin{tabular}{rlrrrrr}
\textbf{Rel.}  &  &
\multicolumn{2}{c}{\textbf{Development Set}} &
\multicolumn{2}{c}{\textbf{Test Set}} \\
\textbf{Size}  & \textbf{Sampling}  &
\textbf{Accuracy} & \textbf{Macro-F1}  &
\textbf{Accuracy} & \textbf{Macro-F1}  &
\textbf{Rep.}  \\
\hline
100\%  & up  &   71.5\% $\pm$    5.7  &   62.5\% $\pm$    4.4  &   71.4\% $\pm$    4.3  &   61.8\% $\pm$    3.0  & 50  \\
\hline
100\%  & none  &   72.2\% $\pm$    3.9  &   58.6\% $\pm$    4.5  &   72.1\% $\pm$    3.2  &   58.9\% $\pm$    3.7  & 50  \\
120\%  & none  &   72.8\% $\pm$    3.1  &   59.6\% $\pm$    4.3  &   72.8\% $\pm$    2.7  &   59.8\% $\pm$    4.3  & 65  \\
140\%  & none  &   73.7\% $\pm$    3.2  &   61.2\% $\pm$    4.1  &   73.5\% $\pm$    2.3  &   61.1\% $\pm$    4.0  & 65  \\
160\%  & none  &   74.4\% $\pm$    3.0  &   62.8\% $\pm$    3.5  &   74.0\% $\pm$    2.2  &   62.3\% $\pm$    3.0  & 65  \\
180\%  & none  &   74.7\% $\pm$    2.8  &   63.3\% $\pm$    3.8  &   74.2\% $\pm$    2.2  &   62.5\% $\pm$    3.5  & 65  \\
200\%  & none  &   75.2\% $\pm$    2.2  &   64.0\% $\pm$    3.0  &   74.5\% $\pm$    1.9  &   63.1\% $\pm$    2.9  & 65  \\
\hline
\end{tabular}
\caption{Development and test set results for \textbf{Llama3 with default ``not harmfull'' label} in scenario 3:
    training a BERT-based classifier on synthetic data matching
    100\% to 200\% of the size available in scenario 1.
    ``Sampling'' refers to the strategy for addressing class
    imbalance in the training data; 
    at least 45 repetitions with different random seeds;}
\label{t:results-s4-r1v1}
\end{table*}

\begin{table*}[ht]
\centering
\begin{tabular}{rlrrrrr}
\textbf{Rel.}  &  &
\multicolumn{2}{c}{\textbf{Development Set}} &
\multicolumn{2}{c}{\textbf{Test Set}} \\
\textbf{Size}  & \textbf{Sampling}  &
\textbf{Accuracy} & \textbf{Macro-F1}  &
\textbf{Accuracy} & \textbf{Macro-F1}  &
\textbf{Rep.}  \\
\hline
100\%  & up  &   72.9\% $\pm$    4.8  &   64.7\% $\pm$    4.0  &   72.7\% $\pm$    3.5  &   64.0\% $\pm$    3.1  & 50  \\
200\%  & up  &   75.1\% $\pm$    3.3  &   68.1\% $\pm$    2.7  &   74.9\% $\pm$    2.8  &   67.5\% $\pm$    2.3  & 45  \\
\hline
100\%  & none  &   73.6\% $\pm$    4.1  &   63.8\% $\pm$    3.8  &   73.4\% $\pm$    3.0  &   63.3\% $\pm$    3.3  & 50  \\
120\%  & none  &   74.8\% $\pm$    3.6  &   65.3\% $\pm$    3.4  &   74.5\% $\pm$    2.5  &   64.7\% $\pm$    3.0  & 65  \\
140\%  & none  &   75.2\% $\pm$    3.4  &   65.9\% $\pm$    3.6  &   75.0\% $\pm$    2.3  &   65.4\% $\pm$    3.0  & 65  \\
160\%  & none  &   76.1\% $\pm$    2.6  &   67.3\% $\pm$    3.1  &   75.3\% $\pm$    2.2  &   66.0\% $\pm$    2.8  & 65  \\
180\%  & none  &   76.0\% $\pm$    2.7  &   67.2\% $\pm$    2.9  &   75.6\% $\pm$    1.9  &   66.3\% $\pm$    2.4  & 65  \\
200\%  & none  &   76.6\% $\pm$    2.5  &   68.3\% $\pm$    2.8  &   75.8\% $\pm$    2.1  &   67.0\% $\pm$    2.4  & 65  \\
\hline
\end{tabular}
\caption{Development and test set results for \textbf{Llama3 with unlabelled messages removed} in scenario 3:
    training a BERT-based classifier on synthetic data matching
    100\% to 200\% of the size available in scenario 1.
    ``Sampling'' refers to the strategy for addressing class
    imbalance in the training data; 
    at least 45 repetitions with different random seeds}
\label{t:results-s4-s1v1}
\end{table*}

\begin{table*}[ht]
\centering
\begin{tabular}{rlrrrrr}
\textbf{Rel.}  &  &
\multicolumn{2}{c}{\textbf{Development Set}} &
\multicolumn{2}{c}{\textbf{Test Set}} \\
\textbf{Size}  & \textbf{Sampling}  &
\textbf{Accuracy} & \textbf{Macro-F1}  &
\textbf{Accuracy} & \textbf{Macro-F1}  &
\textbf{Rep.}  \\
\hline
100\%  & up  &   69.5\% $\pm$    1.8  &   50.7\% $\pm$    3.2  &   71.7\% $\pm$    1.6  &   53.6\% $\pm$    4.8  & 50  \\
\hline
100\%  & none  &   69.7\% $\pm$    0.8  &   44.4\% $\pm$    3.1  &   70.2\% $\pm$    0.9  &   45.1\% $\pm$    4.1  & 50  \\
120\%  & none  &   69.8\% $\pm$    0.7  &   44.3\% $\pm$    2.7  &   70.3\% $\pm$    1.0  &   44.9\% $\pm$    3.9  & 65  \\
140\%  & none  &   69.9\% $\pm$    0.8  &   44.2\% $\pm$    3.0  &   70.2\% $\pm$    1.0  &   44.5\% $\pm$    3.7  & 65  \\
160\%  & none  &   69.9\% $\pm$    0.8  &   44.3\% $\pm$    3.1  &   70.3\% $\pm$    0.9  &   44.9\% $\pm$    3.7  & 65  \\
180\%  & none  &   70.0\% $\pm$    0.7  &   45.1\% $\pm$    3.0  &   70.4\% $\pm$    0.9  &   45.5\% $\pm$    3.9  & 65  \\
200\%  & none  &   70.1\% $\pm$    0.7  &   44.9\% $\pm$    3.0  &   70.4\% $\pm$    0.9  &   45.2\% $\pm$    3.7  & 65  \\
\hline
\end{tabular}
\caption{Development and test set results for \textbf{GPT-4o} in scenario 3:
    training a BERT-based classifier on synthetic data matching
    100\% to 200\% of the size available in scenario 1.
    ``Sampling'' refers to the strategy for addressing class
    imbalance in the training data; at least 45 repetitions with different random seeds
    }
\label{t:results-s4-cg12k}
\end{table*}

\begin{table*}[ht]
\centering
\begin{tabular}{rlrrrrr}
\textbf{Rel.}  &  &
\multicolumn{2}{c}{\textbf{Development Set}} &
\multicolumn{2}{c}{\textbf{Test Set}} \\
\textbf{Size}  & \textbf{Sampling}  &
\textbf{Accuracy} & \textbf{Macro-F1}  &
\textbf{Accuracy} & \textbf{Macro-F1}  &
\textbf{Rep.}  \\
\hline
100\%  & up  &   51.3\% $\pm$    8.2  &   50.7\% $\pm$    8.1  &   53.5\% $\pm$    7.8  &   52.9\% $\pm$    7.5  & 50  \\
\hline
100\%  & none  &   49.9\% $\pm$    8.5  &   49.2\% $\pm$    8.5  &   52.2\% $\pm$    8.0  &   51.6\% $\pm$    7.7  & 50  \\
120\%  & none  &   50.7\% $\pm$    8.6  &   50.0\% $\pm$    8.7  &   52.9\% $\pm$    8.1  &   52.4\% $\pm$    7.9  & 65  \\
140\%  & none  &   50.7\% $\pm$    8.9  &   50.0\% $\pm$    9.0  &   52.9\% $\pm$    8.1  &   52.4\% $\pm$    7.9  & 65  \\
160\%  & none  &   51.2\% $\pm$    8.9  &   50.6\% $\pm$    9.0  &   53.5\% $\pm$    8.1  &   52.9\% $\pm$    7.9  & 65  \\
180\%  & none  &   51.4\% $\pm$    8.8  &   50.8\% $\pm$    8.9  &   53.7\% $\pm$    8.3  &   53.2\% $\pm$    8.1  & 65  \\
200\%  & none  &   52.3\% $\pm$    8.3  &   51.8\% $\pm$    8.3  &   54.2\% $\pm$    7.8  &   53.7\% $\pm$    7.5  & 65  \\
\hline
\end{tabular}
\caption{Development and test set results for \textbf{Grok} in scenario 3:
    training a BERT-based classifier on synthetic data matching
    100\% to 200\% of the size available in scenario 1.
    ``Sampling'' refers to the strategy for addressing class
    imbalance in the training data; at least 45 repetitions with different random seeds 
    }
\label{t:results-s4-gr4k}
\end{table*}

\begin{table*}[ht]
\centering
\begin{tabular}{lrlccccr}
&  &  & \multicolumn{2}{c}{\textbf{Development Set}} & \multicolumn{2}{c}{\textbf{Test Set}} \\
\textbf{Labels} & \textbf{Size} & \textbf{Sampling} & \textbf{Accuracy} & \textbf{Macro-F1} & \textbf{Accuracy} & \textbf{Macro-F1} & \textbf{Rep.} \\
\hline
D0  &   20\%  &  up  &   72.3\% $\pm$    4.3  &   59.2\% $\pm$   12.0  &   71.0\% $\pm$    3.8  &   58.0\% $\pm$   11.2  & 50  \\
D0  &   100\%  &  up  &   78.7\% $\pm$    1.1  &   74.0\% $\pm$    1.4  &   77.9\% $\pm$    0.9  &   72.6\% $\pm$    1.3  & 50  \\
\hline
D0  &   20\%  &  none  &   74.9\% $\pm$    1.9  &   64.0\% $\pm$    5.0  &   73.1\% $\pm$    2.0  &   61.8\% $\pm$    4.8  & 85  \\
D0  &   50\%  &  none  &   78.3\% $\pm$    1.7  &   71.2\% $\pm$    2.6  &   76.4\% $\pm$    1.7  &   68.3\% $\pm$    2.5  & 85  \\
D0  &   80\%  &  none  &   79.7\% $\pm$    1.4  &   73.7\% $\pm$    1.9  &   77.3\% $\pm$    0.9  &   70.0\% $\pm$    1.2  & 85  \\
D0  &   100\%  &  none  &   80.0\% $\pm$    1.2  &   73.9\% $\pm$    1.6  &   77.6\% $\pm$    0.9  &   70.3\% $\pm$    1.3  & 85  \\
D0  &   200\%  &  none  &   80.0\% $\pm$    1.2  &   74.2\% $\pm$    1.6  &   77.5\% $\pm$    0.9  &   70.4\% $\pm$    1.4  & 50  \\
\hline
FU  &   20\%  &  up  &   73.0\% $\pm$    3.5  &   61.0\% $\pm$   12.4  &   72.3\% $\pm$    3.0  &   60.5\% $\pm$   12.1  & 50  \\
FU  &   100\%  &  up  &   79.6\% $\pm$    1.0  &   75.3\% $\pm$    1.1  &   79.2\% $\pm$    0.7  &   74.3\% $\pm$    0.9  & 50  \\
\hline
FU  &   20\%  &  none  &   74.8\% $\pm$    1.9  &   66.9\% $\pm$    2.9  &   73.7\% $\pm$    1.7  &   65.7\% $\pm$    2.5  & 85  \\
FU  &   50\%  &  none  &   79.3\% $\pm$    1.5  &   74.2\% $\pm$    1.9  &   78.0\% $\pm$    1.3  &   72.1\% $\pm$    1.7  & 85  \\
FU  &   80\%  &  none  &   80.1\% $\pm$    0.9  &   75.5\% $\pm$    1.0  &   78.8\% $\pm$    1.0  &   73.2\% $\pm$    1.2  & 85  \\
FU  &   100\%  &  none  &   80.4\% $\pm$    1.0  &   75.9\% $\pm$    1.2  &   79.1\% $\pm$    1.0  &   73.5\% $\pm$    1.2  & 85  \\
FU  &   200\%  &  none  &   80.2\% $\pm$    1.1  &   75.7\% $\pm$    1.2  &   78.8\% $\pm$    0.8  &   73.4\% $\pm$    1.0  & 50  \\
\hline
CH  &   20\%  &  up  &   71.9\% $\pm$    3.0  &   57.1\% $\pm$   10.6  &   72.5\% $\pm$    3.2  &   58.6\% $\pm$   11.6  & 50  \\
CH  &   100\%  &  up  &   77.5\% $\pm$    0.9  &   71.3\% $\pm$    1.3  &   78.3\% $\pm$    0.8  &   72.5\% $\pm$    1.1  & 50  \\
\hline
CH  &   20\%  &  none  &   72.0\% $\pm$    1.7  &   55.3\% $\pm$    6.7  &   72.6\% $\pm$    1.6  &   56.9\% $\pm$    7.1  & 85  \\
CH  &   50\%  &  none  &   75.1\% $\pm$    1.7  &   64.4\% $\pm$    3.7  &   76.4\% $\pm$    1.3  &   66.9\% $\pm$    2.7  & 85  \\
CH  &   80\%  &  none  &   76.2\% $\pm$    1.1  &   67.2\% $\pm$    2.0  &   77.2\% $\pm$    0.9  &   69.0\% $\pm$    1.7  & 85  \\
CH  &   100\%  &  none  &   77.1\% $\pm$    0.9  &   68.7\% $\pm$    1.5  &   77.9\% $\pm$    0.9  &   70.2\% $\pm$    1.4  & 85  \\
CH  &   200\%  &  none  &   77.1\% $\pm$    1.0  &   68.6\% $\pm$    2.1  &   78.0\% $\pm$    0.7  &   70.2\% $\pm$    1.3  & 50  \\
\hline
GR  &   20\%  &  up  &   70.5\% $\pm$    3.4  &   67.3\% $\pm$    3.2  &   69.4\% $\pm$    3.3  &   66.6\% $\pm$    2.6  & 50  \\
GR  &   100\%  &  up  &   75.6\% $\pm$    1.4  &   73.0\% $\pm$    1.4  &   74.6\% $\pm$    1.8  &   71.7\% $\pm$    1.4  & 50  \\
\hline
GR  &   20\%  &  none  &   71.3\% $\pm$    2.6  &   67.3\% $\pm$    2.5  &   70.4\% $\pm$    3.0  &   66.9\% $\pm$    2.2  & 85  \\
GR  &   50\%  &  none  &   75.0\% $\pm$    1.6  &   72.0\% $\pm$    1.7  &   74.2\% $\pm$    2.0  &   71.0\% $\pm$    1.7  & 85  \\
GR  &   80\%  &  none  &   76.1\% $\pm$    1.1  &   73.3\% $\pm$    1.2  &   74.4\% $\pm$    1.6  &   71.3\% $\pm$    1.3  & 85  \\
GR  &   100\%  &  none  &   76.3\% $\pm$    1.2  &   73.7\% $\pm$    1.2  &   75.2\% $\pm$    1.7  &   72.1\% $\pm$    1.4  & 85  \\
GR  &   200\%  &  none  &   76.3\% $\pm$    1.2  &   73.6\% $\pm$    1.2  &   75.1\% $\pm$    1.5  &   72.1\% $\pm$    1.3  & 50  \\
\hline
\end{tabular}
\caption{Development and test set results in scenario~4: training BERT-based classifiers on the training split of the
authentic data with synthetic labels predicted by
(a) D0 = Llama3, assuming ``Not Harmful'' when no label is found in the LLM output,
(b) FU = Llama3, removing dataset items for which no label is found in the LLM output,
(c) CH = ChatGPT and
(d) GR = Grok;
at least 45 repetitions with different random seeds;
also shown for comparison results for training on samples
from 20\% to 80\%, as well as tewo copies (200\%) of the data;
both the training set and the development set are sampled
to the given relative size of the authentic data split;
``Sampling'' refers to the strategy for addressing class imbalance in the training data}
\label{t:results-s4-a}
\end{table*}

\end{document}